\documentclass[letterpaper, 10 pt, conference]{ieeeconf}  
\usepackage[utf8]{inputenc}
\usepackage[T1]{fontenc}
\usepackage[nocompress]{cite}

\makeatletter
\let\NAT@parse\undefined
\makeatother
\usepackage[colorlinks=true,citecolor=blue,linkcolor=black,urlcolor=black,hypertexnames=false]{hyperref}

\IEEEoverridecommandlockouts                              

\usepackage{amsmath}
\usepackage{amssymb}
\usepackage{marvosym}
\usepackage{booktabs}
\usepackage{float}
\usepackage{tikz}
\usetikzlibrary{arrows.meta,backgrounds,calc,fit,positioning}
\usepackage{colortbl}

\definecolor{tabband}{HTML}{EFEFEF}
\definecolor{tabdelta}{HTML}{6C8A5E}
\definecolor{teaserhuman}{HTML}{4F78B8}
\definecolor{teaserrobot}{HTML}{C56A3A}
\definecolor{teaserrepresentation}{HTML}{7B5AA6}
\definecolor{projecturl}{HTML}{1A0DAB}
\newcommand{\hdrmid}[1]{\raisebox{1.45ex}[0pt][0pt]{#1}}

\newcommand{\tabgroup}[1]{%
    \rowcolor{tabband}\multicolumn{7}{@{}l@{}}{%
        \rule[-0.55ex]{0pt}{2.7ex}\hspace{\tabcolsep}\itshape #1}}


\makeatletter
\def\thetable{\@arabic\c@table}
\def\fnum@table{Table~\thetable}
\long\def\@makecaption#1#2{%
\ifx\@captype\@IEEEtablestring%
{\footnotesize #1: #2\par}%
\@IEEEtablecaptionsepspace%
\else
\@IEEEfigurecaptionsepspace%
\setbox\@tempboxa\hbox{\footnotesize #1.~~ #2}%
\ifdim \wd\@tempboxa >\hsize%
\setbox\@tempboxa\hbox{\footnotesize #1.~~ }%
\parbox[t]{\hsize}{\footnotesize \noindent\unhbox\@tempboxa#2}%
\else%
\ifcenterfigcaptions \hbox to\hsize{\footnotesize\hfil\box\@tempboxa\hfil}%
\else \hbox to\hsize{\footnotesize\box\@tempboxa\hfil}%
\fi\fi\fi}
\makeatother

\title{\LARGE \bf
Unified Visual-Tactile-Action Modeling from Human Demonstrations for Dexterous Manipulation
}

\author{%
\authorblockN{%
Wenqiao~Li\textsuperscript{*1}, Qianyou~Zhao\textsuperscript{*3},
Jiawen~Hao\textsuperscript{4}, Xuezhou~Zhu\textsuperscript{3}, Tengyu~Liu\textsuperscript{2},\\
Kaifeng~Zhang\textsuperscript{3}, Chuan~Wen\textsuperscript{\Letter\,1},
Siyuan~Huang\textsuperscript{\Letter\,2}}%
\authorblockA{\footnotesize
\textsuperscript{1}\,Shanghai Jiao Tong University\\
\textsuperscript{2}\,State Key Laboratory of General Artificial Intelligence, BIGAI\\
\textsuperscript{3}\,Sharpa Robotics\\
\textsuperscript{4}\,Beijing Institute of Technology\\
\textsuperscript{*}\,Denotes equal contributions\qquad
\textsuperscript{\Letter}\,Corresponding authors}%
}

\begin{document}

\IEEEaftertitletext{%
    \vspace{-2.0em}%
    \begin{center}
        \centering
        \includegraphics[width=0.84\textwidth]{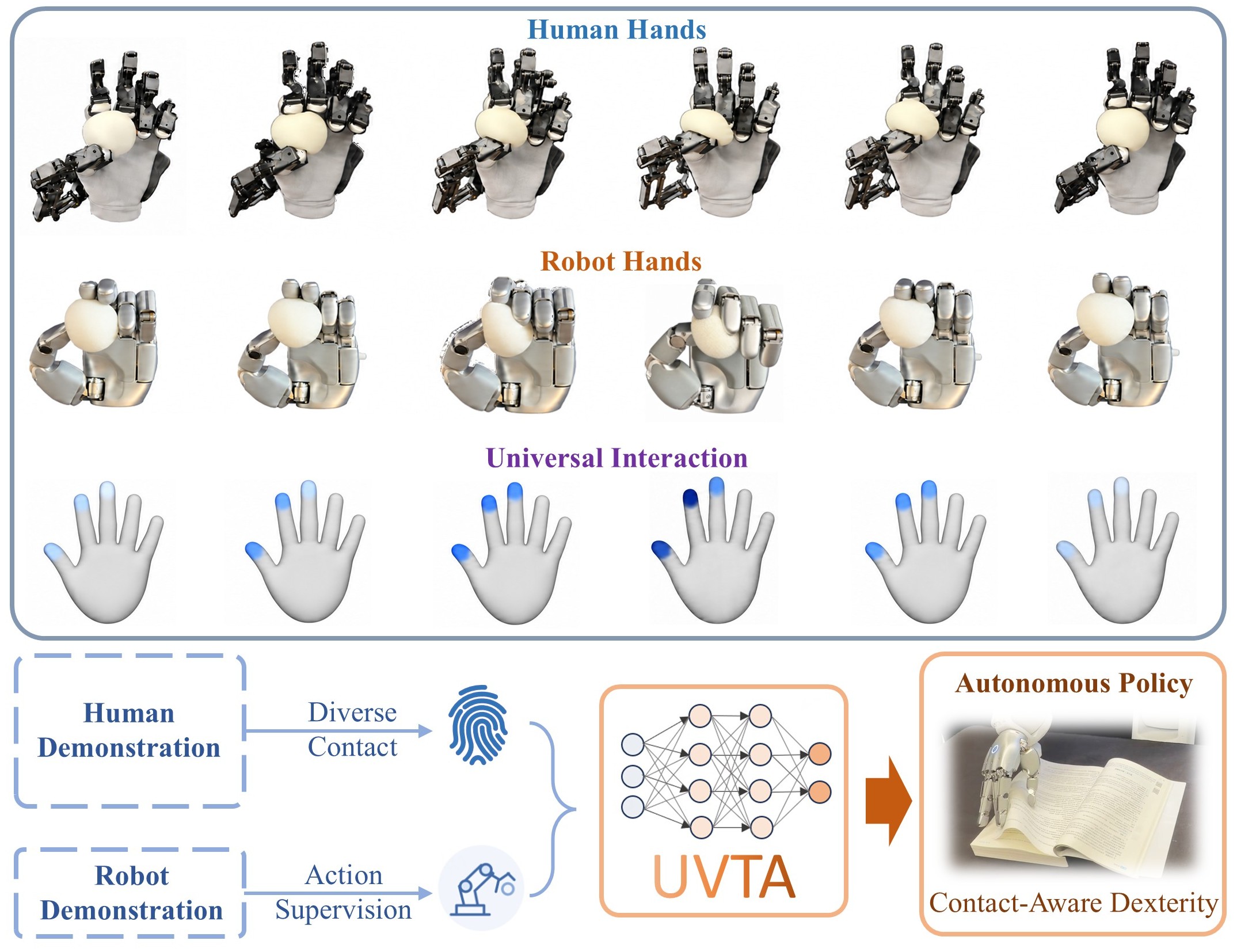}\\[0.35em]
        \refstepcounter{figure}\label{fig:teaser}
        \parbox{\textwidth}{\footnotesize Fig.~\thefigure.~~ We propose Unified Visual Tactile Action (UVTA) modeling, a framework that enables contact-aware dexterity learning from human tactile data. Under the UVTA framework, \textbf{\textcolor{teaserhuman}{Human hand}} and \textbf{\textcolor{teaserrobot}{robot hand}} share a \textbf{\textcolor{teaserrepresentation}{universal interaction representation}.}}
    \end{center}
    \vspace{0.2em}%
}
\maketitle
\thispagestyle{empty}
\pagestyle{empty}

\begin{abstract}
Dexterous manipulation requires tactile feedback. However, robot tactile demonstrations are difficult to scale because dexterous-hand teleoperation provides limited tactile feedback to the operator. In contrast, human demonstrations offer a substantially more scalable source of diverse tactile interactions. Motivated by a simple premise---\emph{hands can change, but the underlying physics of interaction does not}---we leverage human tactile data to improve dexterous manipulation policies. Specifically, we first build a tactile motion-capture system that synchronously records images, tactile signals, and hand motions. Using this system, we construct the UVTA dataset spanning five contact-rich tasks, with 1,000 human demonstrations covering diverse interaction patterns and 150 robot demonstrations per task. To transfer the underlying physics of human interaction to robot control, we propose a Unified Visual-Tactile-Action Model that maps both embodiments into aligned tactile and action representations and jointly predicts future action and tactile trajectories. The joint objective enables human demonstrations to supervise contact-aware representation learning, while only robot actions are executed during deployment. In real-robot evaluations across five tasks, our method achieves an average success rate of 70\%, outperforming the strongest visual-tactile baseline, which achieves 29\%, and an architecture ablation, which achieves 42\%. Performance improves consistently with additional human demonstrations and exhibits no saturation at 1,000 demonstrations per task, validating the effectiveness of scalable human tactile data for dexterous manipulation. Project page is available at {\hypersetup{urlcolor=projecturl}\mbox{\url{https://uni-vta.github.io/}}}.

\end{abstract}


\section{INTRODUCTION}

Tactile feedback is indispensable for dexterous manipulation. Contact-rich tasks depend on contact onset, pressure, deformation, and slip---physical states that are often ambiguous or occluded in visual observations. Access to these states has enabled increasingly precise and reactive manipulation~\cite{tactile_dexterity,see_to_touch,dextac,trex,ttp}. Nevertheless, learning tactile policies typically requires demonstrations collected through dexterous robot-hand teleoperation. Such data are difficult to scale: teleoperating a high-dimensional hand is demanding, and conventional interfaces provide the operator with limited tactile feedback. Consequently, collecting synchronized visual, tactile, and action trajectories remains costly, constraining the diversity of objects, environments, and contact patterns represented in robot datasets.

Human demonstrations provide a substantially more scalable source of diverse physical interactions. Large-scale human video has supported visual representation learning and long-horizon robot control~\cite{ego4d,hoi4d,mvp,r3m,mimicplay}, while motion reconstruction, retargeting, and human--robot co-training transfer human behavior more directly~\cite{dexmv,dexcap,egomimic,deximit,egovla,egoscale}. However, human and robot hands differ in morphology, kinematics, sensing layouts, and action spaces, making direct motion correspondence unreliable. Moreover, video alone does not fully capture the local contact dynamics that determine manipulation success. We therefore seek to transfer human interaction experience through a quantity that is more fundamental than embodiment-specific motion. Our central premise is that \emph{hands can change, but the underlying physics of interaction does not}: although embodiments differ, contact formation, deformation, slip, release, and object response retain shared physical structure.

To capture this structure at scale, we first develop a wearable tactile motion-capture system that synchronously records wrist-centric images, fingertip tactile measurements, and hand motion. We then use the system to construct the UVTA dataset over five contact-rich tasks. Each task contains 1,000 human demonstrations that cover diverse interaction patterns and 150 robot demonstrations. This composition combines the diversity and scalability of human interaction with the embodiment-specific supervision required for robot execution.

Building on this dataset, we propose a \emph{Unified Visual-Tactile-Action Model} that maps human and robot trajectories into aligned tactile and action representations. The model jointly predicts future action and tactile trajectories, using human demonstrations to supervise contact-aware representation learning while executing only robot actions at deployment. Predicting tactile outcomes connects motion to its physical consequences, providing a shared learning signal across embodiments~\cite{unipi,gr1,robodreamer,vpp,uva,flare,dreamdojo,wam_zero,egowam,imagine2touch,touchworld,vitacformer,vtam}. Across five real-robot tasks, our method achieves a 70\% average success rate, compared with 29\% for the strongest visual-tactile baseline and 42\% for the ablated architecture. Performance continues to improve as the number of human demonstrations increases, with no saturation at 1,000 demonstrations per task. These results demonstrate that scalable human tactile data can substantially improve contact-rich dexterous manipulation.

Our main contributions are threefold:
\begin{itemize}
    \item We develop a wearable tactile motion-capture system that synchronously captures visual observations, fingertip tactile measurements, and hand motion without requiring robot teleoperation.
    \item We introduce the UVTA dataset, comprising five contact-rich tasks with 1,000 diverse human demonstrations and 150 robot demonstrations per task.
    \item We propose a unified visual-tactile-action model that transfers contact-aware physical representations across embodiments. Real-robot experiments and scaling studies validate both its performance and its ability to benefit from increasing amounts of human tactile data.
\end{itemize}

\begin{figure*}[!t]
    \centering
    \includegraphics[width=\textwidth]{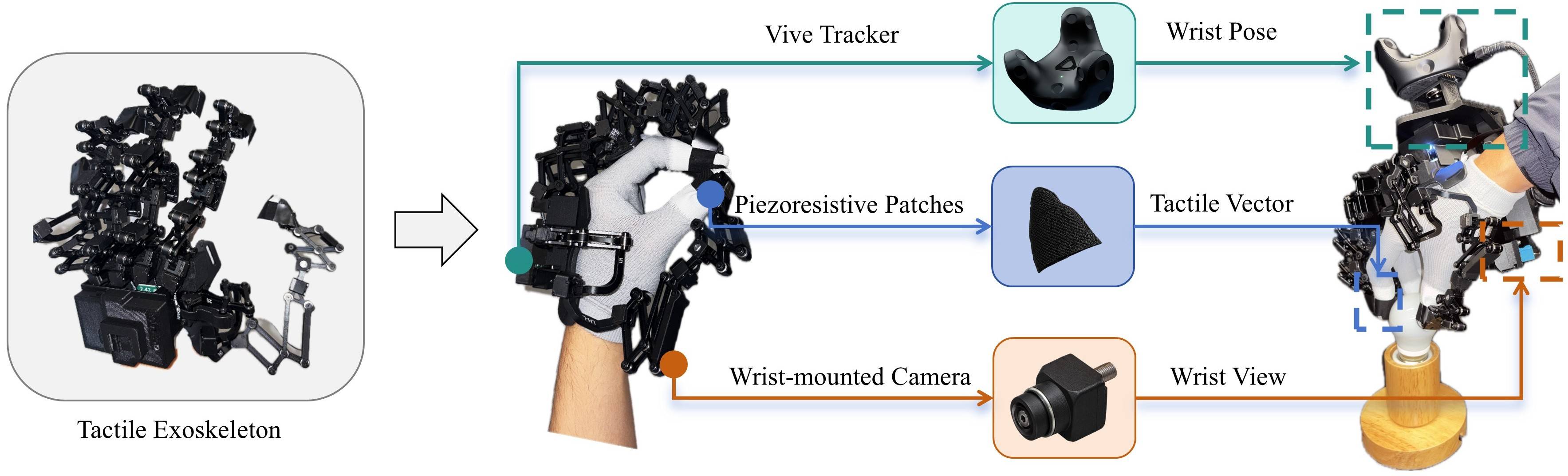}
    \caption{\textbf{Overview of the Tactile Motion-Capture System.} A passive tactile exoskeleton captures hand motion and tactile signals. The integrated system further includes a VIVE tracker and a wrist-mounted RGB camera for capturing the wrist pose and wrist-centric images, respectively. The camera and VIVE tracker are calibrated to the Sharpa Wave wrist frame.}
    \label{fig:human_data_system}
\end{figure*}

\begin{figure}[!t]
    \centering
    \includegraphics[width=\columnwidth]{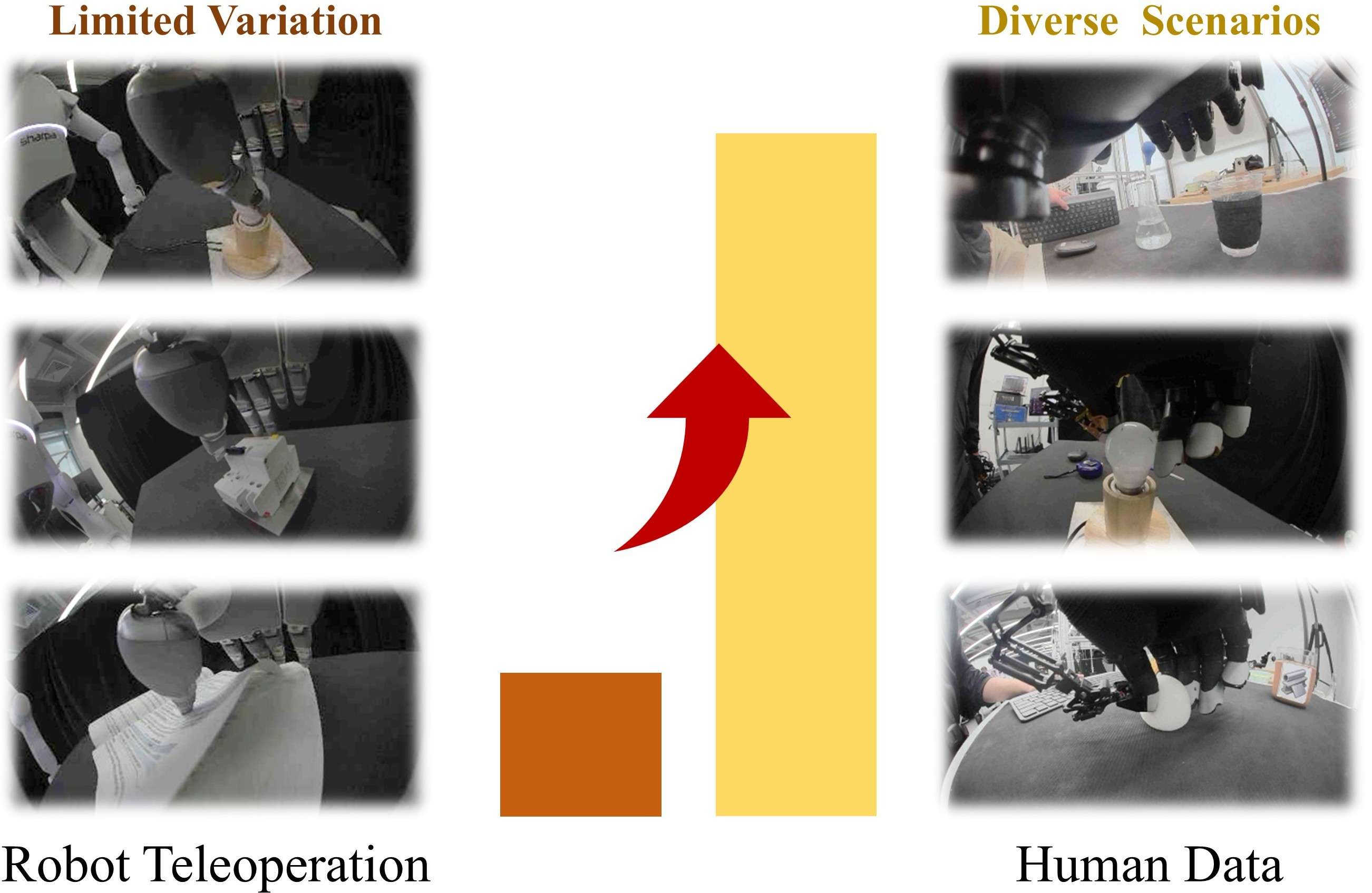}
    \caption{\textbf{Dataset Overview.} Each task contains 150 robot demonstrations collected under limited variation, and 1,000 human demonstrations with diverse scenarios and contact patterns.}
    \label{fig:dataset}
\end{figure}

\begin{figure*}[!t]
    \centering
    \includegraphics[width=\textwidth]{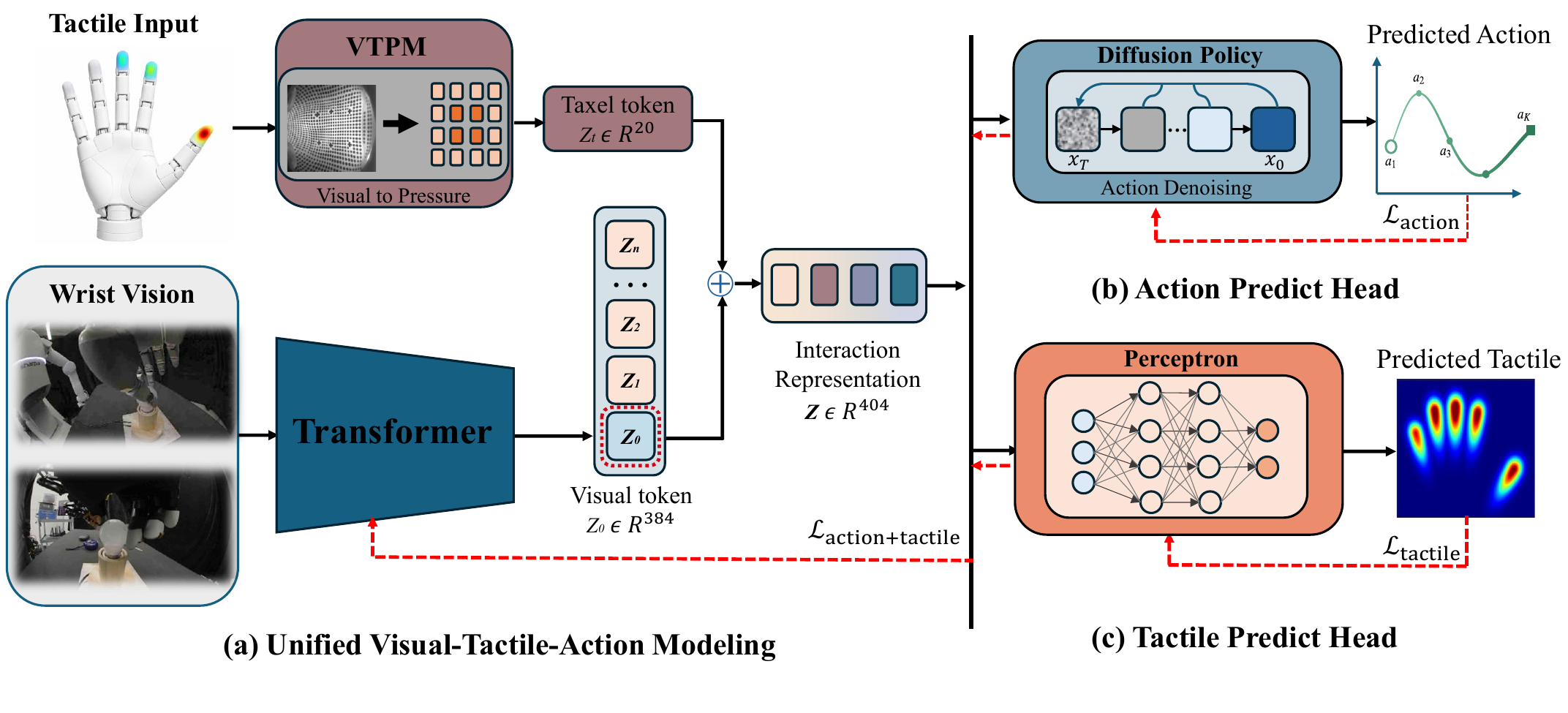}
    \caption{\textbf{Overview of Our Unified Visual-Tactile-Action Model.} VTPM converts the visual-tactile input into a 20-D piezoresistive taxel token, while a Transformer extracts a 384-D visual token. The two tokens are concatenated into a unified 404-D interaction representation and passed to a diffusion action head and a tactile regression head. The action and tactile losses supervise their respective prediction heads, while their joint objective optimizes the shared Transformer to learn a unified tactile-aware visual representation.}
    \label{fig:method_overview}
\end{figure*}

\section{Related Work}

\textbf{Robot Learning from Human Data.}
Large-scale egocentric datasets capture diverse human interactions across tasks, objects, and environments~\cite{ego4d,hoi4d,epic_kitchens_100,ego_exo4d}. This experience has been used to learn reusable visual representations and latent plans for downstream robot control~\cite{mvp,r3m,vip,vc1,mimicplay}, while a complementary line transfers human motor behavior through pose reconstruction, action retargeting, portable motion capture, or human--robot co-training~\cite{dexmv,dexcap,egomimic,deximit,egovla,egoscale,humanego}. Although these methods substantially reduce the demand for robot demonstrations, direct policy transfer still relies on recoverable human motion or an aligned action representation. Predictive approaches instead use embodiment-invariant interaction outcomes as supervision~\cite{egowam}. Our work jointly predicts action and future tactile trajectories from both human and robot demonstrations. Human actions contribute training supervision but are not treated as robot commands; deployment selects only robot-action predictions, while paired action--touch dynamics provide the shared cross-embodiment representation.

\textbf{Tactile Learning for Dexterous Manipulation.}
Touch reveals contact, force, slip, and deformation that may be ambiguous or occluded in vision, enabling closed-loop dexterity in contact-rich tasks~\cite{feeling_success,making_sense,tactile_dexterity,see_to_touch,dextac}. Beyond task-specific control, tactile representations have been learned from robotic play, masked multimodal objectives, large unlabeled tactile corpora, and heterogeneous sensor data~\cite{touch_and_go,tdex,power_senses,sparsh,anytouch}. Policy-level systems further combine vision and touch for bimanual or multifingered manipulation, high-frequency reactive control, and physically grounded vision-language-action learning~\cite{hato,vitacformer,rdp,tactile_vla,vtam}. Recent scaling efforts introduce tactile mid-training and human-centric tactile pre-training~\cite{trex,ttp}. Nevertheless, existing systems predominantly learn from robot touch or transfer human touch through staged training with aligned human--robot action spaces. We instead co-train human tactile motion-capture and robot teleoperation trajectories with the same action and future-tactile objectives. Their coupling lets both embodiments shape the visual representation, while deployment retains robot-specific execution.

\textbf{World Models for Robot Learning.}
World models support robot learning by predicting future observations or latent states and coupling these predictions to planning or action generation~\cite{world_models,planet,daydreamer,tdmpc2,unipi,gr1,robodreamer}. Recent systems integrate future-video or future-feature prediction directly into generalist policies and jointly model visual dynamics and actions~\cite{vpp,uva,flare}. Scaling this paradigm with heterogeneous robot and human videos has produced world action models that transfer physical knowledge across tasks and embodiments~\cite{dreamdojo,wam_zero,egowam}. However, most such models predict visual futures, in which contact onset, force, and local deformation are only weakly observable. Tactile prediction offers a complementary world-modeling target, from anticipating local touch observations to generating tactile subgoals and multimodal interaction dynamics~\cite{touch_and_go,imagine2touch,touchworld,vtam,ttp}. Our model jointly predicts future tactile and action trajectories from both human and robot interactions, coupling physical consequences with the motions that produce them. At deployment, only the robot-action predictions are executed.\looseness=-1

\section{Data Collection System and Dataset}

We collect human demonstrations rich in contact using the wearable tactile exoskeleton glove system shown in Fig.~\ref{fig:human_data_system}. The system records wrist fisheye images, fingertip tactile signals, and 22-DoF joint angles. After processing and temporal synchronization, the raw signals captured by this system are aligned with Sharpa Wave Hand in both kinematic and contact space.

\subsection{Tactile Motion-Capture System}

\textbf{Tactile sensing.}
We employ conventional force-sensing resistors (FSR) to capture human tactile signals. Specifically, each fingertip of the tactile-sensing exoskeleton glove is equipped with an FSR array containing four sensing regions, each with an active area of $15\times15 \mathrm{mm}^2$. When external pressure is applied to the fingertip, the resistance of the corresponding sensing point changes. This change is converted into an electrical signal and recorded as the tactile measurement at that location.

\textbf{Hand Motion.}
Wrist and finger motions are captured using a pair of VIVE Tracker 3.0 devices and a passive linkage exoskeleton, respectively. One tracker is fixed to the table to establish a reference frame, whereas the other is rigidly attached to the wrist via a 3D-printed adapter. At each timestep, the pose of the wrist tracker relative to the reference tracker is composed with the calibrated tracker-to-wrist transformation to recover the 6-DoF wrist pose. Finger motion is measured by a passive 22-DoF linkage exoskeleton mounted on the dorsum of the hand. The exoskeleton combines rigid carbon-fiber links with a compliant dorsal support plate. Forward kinematics converts the encoder measurements into the positions and orientations of the five fingertips, which are then transformed into the Sharpa Wave palm frame using a fixed calibration. A constrained inverse-kinematics optimization recovers the corresponding 22-DoF hand configuration by matching fingertip poses while preserving the relative geometry between the thumb and the other fingertips. Joint limits are enforced as hard constraints, whereas the thumb carpometacarpal (CMC) and four-finger metacarpophalangeal (MCP) couplings are modeled as soft linear constraints.

\textbf{Image Stream.}
A monocular fisheye RGB camera is rigidly mounted near the wrist using a fixed adapter. Its camera-to-wrist extrinsic transformation is calibrated to match the camera-to-end-effector transformation of the Sharpa North robot, thereby maintaining consistent camera--hand geometry across embodiments. The wrist image stream, tactile signals, wrist poses, and joint motion are timestamped using a shared host clock and temporally resampled at 30 Hz.


\subsection{Dataset}

Our dataset covers five contact-rich manipulation tasks, each with 150 robot and 1,000 human demonstrations. Robot demonstrations are collected through teleoperation on the Sharpa North platform, with the scene layout, initial robot configuration, and table height fixed for each task. Human demonstrations are collected using our wearable tactile motion-capture system without a robot in the loop. We vary object instances and scene backgrounds, and use black and white gloves to vary hand appearance. Fig.~\ref{fig:dataset} contrasts the fixed robot collection settings with the more varied human demonstrations.

Robot demonstrations pair visual-tactile observations with executable wrist and finger commands, providing direct supervision for control on the target platform.
Human demonstrations contribute more varied motion and contact sequences for learning contact-aware representations across objects and scenes. Both sources supervise action and future-tactile prediction, training the shared representation to capture how contact changes with motion. Human motions serve as training targets rather than directly executable robot commands.

\section{Method}

We aim to learn generalizable interaction representations from human tactile demonstrations for dexterous manipulation. To bridge the visual, tactile, and action embodiment gaps between human and robot data, we align both into a unified cross-embodiment space (Sec.~\ref{sec:unified_space}). As shown in Fig.~\ref{fig:method_overview}, our Unified Visual-Tactile-Action Model conditions action generation and future-tactile prediction on the same wrist-centric visual-tactile observation (Sec.~\ref{sec:model_architecture}). It is trained jointly on human motion-capture and robot teleoperation samples using a balanced co-training recipe; during inference, only robot-normalized action predictions are converted into wrist and finger commands.

\subsection{Problem Formulation}
We consider contact-rich dexterous manipulation with two data sources: a human dataset $\mathcal{D}^{\mathrm{H}}$ collected by our tactile motion-capture system and a robot dataset $\mathcal{D}^{\mathrm{R}}$ collected by teleoperation. A sample from either embodiment contains, at each timestep, a synchronized wrist RGB image $\mathbf{I}_{t}$, a tactile token $Z_{t}\in\mathbb{R}^{20}$, and an action $\mathbf{a}_{t}\in\mathbb{R}^{31}$. Given the current visual-tactile observation, the model predicts a future action chunk and a future tactile trajectory over horizon $H$:
\begin{equation}
    \mathbf{A}_{t}=\mathbf{a}_{t:t+H-1},\qquad
    Y_{t}=Z_{t+1:t+H},
\end{equation}
where $H=16$ in our implementation. Robot actions are executable teleoperation commands, whereas human actions are derived from consecutive motion-capture states. We formulate policy learning as the joint prediction of future human and robot interaction trajectories:
\begin{equation}
    (\widehat{\mathbf{A}}_{t},\widehat{Y}_{t})
    \sim \pi_{\Theta}(\cdot\mid \mathbf{I}_{t},Z_{t}),
\end{equation}
where $\Theta=\{\theta,\phi,\psi\}$ denotes the parameters of the visual encoder, action head, and tactile head, respectively. At test time, only the robot action prediction $\widehat{\mathbf{A}}_{t}$ is sent to the controller.

\subsection{Unified Action and Tactile Space}
\label{sec:unified_space}
We aim to learn general physical representations from human tactile demonstrations that are directly useful for dexterous manipulation. However, human tactile motion-capture data differs from robot data in both tactile representation and kinematics, while inevitably containing tactile and motion noise.
To align robot and human tactile observations, we introduce Visual-Tactile-to-Piezoresistive Taxel Mapping (VTPM), which converts visual-tactile deformation maps captured by the Wave sensors into pressure-like piezoresistive taxel signals. For each finger, valid pixels are assigned to four glove-aligned regions in a canonical fingertip frame, and the deformation intensity within each region is aggregated into one scalar value. Concatenating all five fingers produces a unified 20-dimensional tactile representation shared by human and robot demonstrations. For the action space, human and robot motions are represented using the same 31-dimensional layout. Wrist motion is expressed relative to the current wrist frame using 3-D translation and a continuous 6-D rotation representation, while finger motion is represented by 22 absolute joint angles. This relative-wrist and absolute-finger formulation reduces dependence on the global workspace while preserving precise hand configurations for dexterous manipulation.

\subsection{Model Architecture}
\label{sec:model_architecture}
As shown in Fig.~\ref{fig:method_overview}, the model has a shared visual-tactile condition and two output branches. We omit the embodiment superscript on intermediate features for clarity. A trainable transformer encoder $f_{\theta}$ maps the current $224\times224$ wrist image to a 384-D visual token,
\begin{equation}
    Z_{0}=f_{\theta}(\mathbf{I}_{t})\in\mathbb{R}^{384}.
\end{equation}
The encoder uses a ViT-S/8 architecture and is optimized from scratch in our implementation. In parallel, VTPM provides the 20-D tactile token $Z_t$, after normalization with statistics specific to embodiment $e$. The model performs early feature-level fusion by direct concatenation,
\begin{equation}
    Z=
    [Z_{0};Z_{t}]
    \in\mathbb{R}^{404}.
\end{equation}

The action branch models a 16-step, 31-D action chunk with conditional diffusion. Let $\mathbf{x}_{0}=\mathcal{N}_{e}(\mathbf{A}_{t})\in\mathbb{R}^{16\times31}$, where $e\in\{\mathrm{H},\mathrm{R}\}$ selects the normalization statistics of the sample's embodiment, and each step contains a 9-D relative wrist target (3-D translation and 6-D rotation) and 22 absolute finger-joint targets. At diffusion step $k$, Gaussian noise $\boldsymbol{\epsilon}\sim\mathcal{N}(\mathbf{0},\mathbf{I})$ is added as
\begin{equation}
    \mathbf{x}_{k}=\sqrt{\bar{\alpha}_{k}}\mathbf{x}_{0}
    +\sqrt{1-\bar{\alpha}_{k}}\boldsymbol{\epsilon}.
\end{equation}
A FiLM-conditioned 1-D U-Net $\epsilon_{\phi}(\mathbf{x}_{k},k,Z)$ predicts the injected noise. Its channel widths are 256, 512, and 1024. In parallel, the tactile world head is an MLP comprising two linear layers: the first maps the 404-D interaction representation to a 512-D hidden feature and applies a Mish activation, while the second produces a 320-D output. It directly regresses the complete future tactile trajectory from the noise-free condition:
\begin{equation}
    \widehat{Y}_{t}=g_{\psi}(Z)
    \in\mathbb{R}^{16\times20}.
\end{equation}
The tactile target is shifted one step ahead and aligned with the action horizon. Keeping it outside the diffusion trajectory prevents tactile denoising errors from entering the iterative action-generation process, while its regression loss still updates the shared visual encoder through $Z$.

\subsection{Training Recipe}
We train each task-specific model in a single human--robot co-training run, rather than using sequential pretraining, mid-training, and post-training stages. The training set is the union $\mathcal{D}^{\mathrm{H}}\cup\mathcal{D}^{\mathrm{R}}$. A weighted random sampler assigns equal probability mass to the human and robot datasets and samples with replacement, preventing their different numbers of frames from determining the mixture ratio.

We apply min--max normalization separately to each embodiment so that both human and robot channels occupy the same numerical range; the saved robot statistics are used to decode predictions at deployment. To deal with tactile sensor native noise problem, tactile signals are baseline-corrected per recording before normalization. For human data, the baseline is the mean of the first five frames; for robot data, it is the first frame. 

For each sample, we draw a diffusion step $k$ uniformly from 50 training noise levels. The denoising objective averages the wrist and hand errors separately and weights the two groups equally, so the 22-D hand block does not dominate the 9-D wrist block:
\begin{equation}
    \mathcal{L}_{\mathrm{action}}=
    \tfrac{1}{2}\operatorname{MSE}
    (\boldsymbol{\epsilon}_{\mathrm{wrist}},\widehat{\boldsymbol{\epsilon}}_{\mathrm{wrist}})
    +\tfrac{1}{2}\operatorname{MSE}
    (\boldsymbol{\epsilon}_{\mathrm{hand}},\widehat{\boldsymbol{\epsilon}}_{\mathrm{hand}}),
\end{equation}
where each $\operatorname{MSE}$ is itself averaged over batch, time, and dimensions within that group. The future-tactile loss is
\begin{equation}
    \mathcal{L}_{\mathrm{tactile}}=
    \operatorname{MSE}\!\left(
    g_{\psi}(Z),\mathcal{N}_{e}(Y_{t})
    \right),
\end{equation}
and the optimized objective is
\begin{equation}
    \mathcal{L}_{\mathrm{action+tactile}}
    =\mathcal{L}_{\mathrm{action}}
    +\lambda\mathcal{L}_{\mathrm{tactile}}.
\end{equation}
The tactile-loss weight $\lambda$ is set to 0.2 in all experiments. We optimize all modules jointly for 300 epochs with batches of 250 using AdamW.

During rollout, DDIM uses 16 reverse steps to sample a 16-step robot action chunk. We invert the robot normalization, convert the 6-D rotations back to $\mathrm{SO}(3)$, and anchor the relative predictions of the wrist to the live pose as $\widehat{\mathbf{T}}_{t+k}=\mathbf{T}_{t}\Delta\widehat{\mathbf{T}}_{t,k}$. The controller executes the first eight wrist and hand waypoints before replacing the newly visual-tactile feedback. The tactile-head output is not used for control.

\section{Experiments}

\begin{table*}[t]
    \centering
    \caption{\textbf{Comparison of Our Method with Baseline and Ablation Methods Across Five Tasks.} Success rates (\%) are computed over 10 rollouts per task, then averaged across tasks. The last row is our full model; the blocks above compare it against representative baselines and against ablations of human data only, future-tactile prediction, tactile input, and proprioceptive conditioning.}
    \label{tab:main_results}
    \setlength{\tabcolsep}{7pt}
    \renewcommand{\arraystretch}{1.0}
    \begin{tabular*}{\textwidth}{@{\extracolsep{\fill}}l|ccccc!{\hspace{7pt}}|!{\extracolsep{0pt}}c}
        \toprule
        \hdrmid{\textbf{Configuration}} & \shortstack{Flip\\Page} & \shortstack{Light\\Bulb}
            & \shortstack{Toggle\\Switch} & \shortstack{Ball\\Classification}
            & \shortstack{Liquid\\Transfer}
            & \hdrmid{\makebox[1.75cm]{\textbf{Average}}} \\
        \midrule
        \tabgroup{Baseline Methods} \\
        ViTacFormer~\cite{vitacformer} & 0 & 4 & 0 & 0 & 0 & 1 \\
        RDP~\cite{rdp}                 & 10 & 20 & 20 & 18 & 0 & 14 \\
        T-Rex~\cite{trex}              & 43 & 26 & 24 & 36 & 15 & 29 \\
        \midrule
        \tabgroup{Ablation Methods} \\
        Human Data only         & 0 & 0 & 0 & 0 & 0 & 0 \\
        w/o Tactile Prediction  & 58 & 38 & 50 & 44 & 21 & 42 \\
        Vision-only             & 47 & 23 & 26 & 20 & 14 & 26 \\
        w/ Proprioception       & 22 & 20 & 14 & 20 & 15 & 18 \\
        \midrule
        \textbf{Ours} & \textbf{83} & \textbf{80} & \textbf{88}
            & \textbf{56} & \textbf{44} & \textbf{70} \\
        \bottomrule
    \end{tabular*}
\end{table*}

\begin{figure*}[t]
    \centering
    \includegraphics[width=\textwidth]{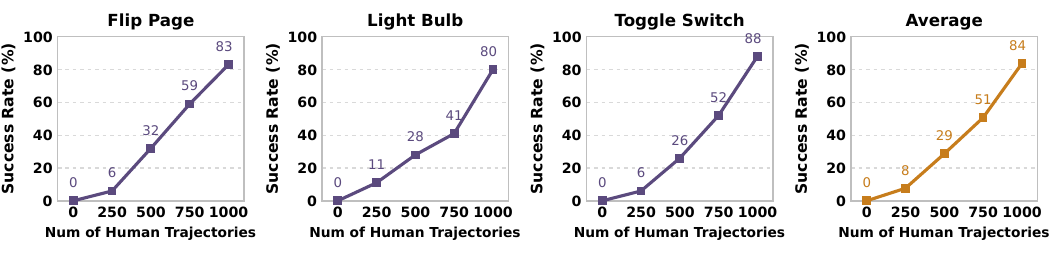}
    \caption{\textbf{Human Tactile Data Scaling Curve.} Success rates on three contact-rich tasks and their average as we add human demonstrations to a fixed budget of robot demonstrations. Performance keeps improving with more human data and has not saturated at 1,000 trajectories.}
    \label{fig:human_scaling}
\end{figure*}

In this section, we aim to answer the following questions through our experiments:\\ 
\textbf{RQ1:} \textit{What is the value of human tactile data for contact-rich dexterous manipulation?} \\
\textbf{RQ2:} \textit{Does scaling human tactile data improve real-robot performance?} \\ 
\textbf{RQ3:} \textit{How important are the architecture and training design of the world model?} 
\subsection{Experiment Setup}
\textbf{Tasks:} To evaluate policy performance, we design five contact-rich manipulation tasks.
Each task is provided with 150 robot demonstrations and 1,000 human demonstrations.\\
\textbf{(Task I)}  \textit{Flip Page.}
The robot approaches the book, slows down near the page, and coordinates its fingers to rub and separate a single page before turning it to the left.\\
\textbf{(Task II)} \textit{Screw Light Bulb.}
The robot approaches a light bulb loosely placed in its socket and coordinates its wrist and fingers to screw the bulb into the socket until it lights up.\\
\textbf{(Task III)} \textit{Toggle Switch.}
The robot approaches a circuit breaker on the table, stabilizes it with the index and middle fingers, and uses the thumb to firmly push the switch until it is fully turned on.\\
\textbf{(Task IV)} \textit{Ball Classification.}
The robot approaches visually similar balls made of different materials, infers their material properties through tactile interaction, and sorts them into the corresponding cups.\\
\textbf{(Task V)} \textit{Liquid Transfer with a Dropper.}
The robot grasps a dropper, inserts it into a flask to draw water, transfers it to a target cup, and dispenses the water without spilling.\\
\textbf{Robot Platform and Action Space:} All real-world experiments use
a fixed-base North robot with two 22-DoF Sharpa Wave dexterous hands.
The policy receives the RGB images from the right-arm monocular wrist camera and calibrated tactile readings from the fingertips. Actions use relative end-effector delta control for the right arm and absolute joint control for the right-hand fingers.\\
\textbf{Baselines:} We compare our method with three representative visuo-tactile policies:
(1) \textbf{ViTacFormer}~\cite{vitacformer}, an ACT-style dexterous imitation policy that learns cross-modal visual-tactile representations through cross-attention and auxiliary future tactile prediction; (2) \textbf{Reactive Diffusion Policy (RDP)}~\cite{rdp}, a slow-fast diffusion policy that combines low-frequency visual action-chunk generation with high-frequency tactile-reactive refinement for contact-rich manipulation; and (3) \textbf{T-Rex}~\cite{trex}, a pretrained tactile-reactive dexterous foundation model that uses a variable-rate Mixture-of-Transformers to couple low-frequency visuo-motor planning with high-frequency tactile action refinement.

For a fair comparison, all methods receive the same wrist RGB and per-finger tactile observations, predict the same right-arm and right-hand action space, and use identical task-specific human and robot demonstration splits and evaluation protocols. Each baseline retains its original tactile fusion and policy architecture and is trained or post-trained on the same number of robot demonstrations as our method.\\
\textbf{Evaluation Metric:} We evaluate each method and ablation over 10 real-robot trials per task, with randomized object positions and heights. Rather than scoring each rollout as a binary outcome, we report a stage-wise success rate that awards partial credit for reaching intermediate stages. For example, in the flip page task, a rollout receives a score of 0.3 for successfully separating the pages, 0.7 for turning multiple pages, and 1.0 for successfully turning exactly one page.

\subsection{Main Results and Ablations}
Table~\ref{tab:main_results} jointly reports the baseline comparison and controlled ablations. Scores are averaged over 10 real-robot rollouts and reported as percentages. The final column is the unweighted mean over the five tasks.

\textbf{Human Tactile Data is Key to Strong Contact-Rich Manipulation Policy.}
Our method achieves a 70\% average success rate and performs best on all five tasks, exceeding the strongest baseline, T-Rex, by 41 percentage points. The gains are particularly pronounced on \textit{Toggle Switch} (88\% versus 24\%) and \textit{Screw Light Bulb} (80\% versus 26\%), where successful execution depends on contact events that are difficult to resolve from wrist vision alone. In contrast, training with human data alone yields 0\% because human trajectories are not directly executable by the robot. Together with the 0\% result obtained with no human trajectories in Fig.~\ref{fig:human_scaling}, these results show that neither source is sufficient in isolation under the available data budget. The improvement therefore comes from combining diverse human interaction experience with robot-specific action supervision through unified co-training.

\textbf{Performance scales with human tactile data and has not saturated.}
Fig.~\ref{fig:human_scaling} varies the number of human trajectories while keeping the 150 robot demonstrations per task fixed. Averaged over \textit{Flip Page}, \textit{Screw Light Bulb}, and \textit{Toggle Switch}, success increases monotonically from 0\% without human data to 8\%, 29\%, 51\%, and 84\% with 250, 500, 750, and 1{,}000 human trajectories, respectively. All three tasks exhibit the same positive trend, indicating that the benefit is not confined to a single manipulation behavior. Moreover, the largest average gain, 33 percentage points, occurs between 750 and 1{,}000 trajectories. The curve thus remains steep at the largest dataset size evaluated, suggesting that performance has not yet saturated and may continue to benefit from additional human tactile demonstrations.

\textbf{Future Tactile Prediction is Critical to the Model.}
The ablations in Table~\ref{tab:main_results} isolate the contributions of tactile conditioning and future-tactile prediction. Removing the tactile prediction head while retaining tactile input reduces the average success rate from 70\% to 42\%, demonstrating that future tactile supervision is more than an auxiliary output: it encourages the shared representation to encode the physical consequences of actions. Removing tactile input as well further reduces performance to 26\%, confirming that current contact observations provide information unavailable from vision alone. Finally, adding proprioceptive conditioning produces the largest degradation, reaching only 18\%. We observe that the proprioception-conditioned policy relies heavily on its current state when predicting actions and consequently produces behaviors resembling the replay of a nearly fixed demonstration trajectory. This reliance limits its ability to adapt actions in response to changing visual and tactile observations, supporting our choice to condition both prediction heads on the unified visual-tactile representation.

\section{Conclusion}
We presented UVTA, a framework for leveraging scalable human tactile demonstrations in contact-rich dexterous manipulation. Motivated by the premise that hand embodiments may differ while the underlying physics of interaction remains shared, we developed a wearable tactile motion-capture system that synchronously records wrist-centric vision, fingertip tactile measurements, and hand motion. Using this system, we collected a dataset spanning five tasks, with 1,000 diverse human demonstrations and 150 robot demonstrations per task. Our Unified Visual-Tactile-Action Model aligns the two embodiments and jointly predicts future action and tactile trajectories, allowing human data to supervise contact-aware representation learning while retaining robot-specific execution. Across five real-robot tasks, the resulting policy achieves a 70\% average success rate, compared with 29\% for the strongest visual-tactile baseline. Ablations further show that removing future-tactile prediction reduces performance to 42\%, confirming the importance of modeling the physical consequences of actions. Finally, performance improves consistently as human data scales and remains unsaturated at 1,000 demonstrations per task. These results establish human tactile interaction as a practical and scalable source of physical supervision for dexterous robot learning.

\IEEEtriggeratref{31}
\bibliographystyle{ieeetr}
\bibliography{references}

\clearpage
\appendices
\small
\raggedbottom

\section{Additional System and Data Details}
\label{app:system_data}

\subsection{Cross-Embodiment Calibration}
The human and robot streams are aligned at both the observation and action levels. For vision, the wrist-camera adapter on the motion-capture system is calibrated so that its camera-to-wrist transformation matches the camera-to-end-effector transformation used on the robot. For motion, one VIVE tracker defines the table reference frame and a second tracker measures the human wrist pose. The calibrated tracker-to-wrist transformation converts these measurements into wrist poses, while constrained inverse kinematics maps the passive exoskeleton measurements to the 22 joint angles of the Sharpa Wave hand. Wrist targets are then expressed relative to the current wrist frame using 3-D translation and a continuous 6-D rotation representation. This construction avoids requiring the human and robot to share a global workspace.

\subsection{Tactile Processing and Synchronization}
Each human fingertip contains four FSR sensing regions. The five fingertips therefore produce the same 20-D tactile layout used by VTPM for robot observations. All image, tactile, wrist-pose, and hand-motion streams are timestamped with a shared host clock and resampled to 30~Hz. Before normalization, tactile readings are baseline corrected independently for each recording. We subtract the mean of the first five frames for human recordings and the first frame for robot recordings, reflecting the different resting-noise characteristics of the two sensors. Human and robot tactile channels are subsequently normalized using embodiment-specific statistics.

\subsection{Dataset Composition}
The complete dataset contains 5,750 demonstrations across five tasks: 5,000 human demonstrations and 750 robot demonstrations. Robot data provide executable commands on the target platform, whereas human data expand the variation in object instances, backgrounds, hand appearance, motion, and contact patterns. Every training sample contains a synchronized wrist image, a 20-D tactile token, and a 31-D motion target. Human targets supervise representation learning in the unified action space but are never issued directly as robot commands.

\section{Implementation Details}
\label{app:implementation}

Table~\ref{tab:implementation_details} summarizes the model and optimization settings used in all experiments. We train one policy per task. The weighted sampler assigns equal probability mass to the human and robot datasets and samples with replacement, preventing the larger human dataset from determining the training mixture. Min--max statistics are computed separately for the two embodiments; at deployment, only the robot statistics are used to decode executable actions. Exponential moving average weights are used for inference.

\begin{table}[!t]
    \centering
    \caption{\textbf{Implementation Details.} Settings shared by all task-specific UVTA policies.}
    \label{tab:implementation_details}
    \setlength{\tabcolsep}{4pt}
    \renewcommand{\arraystretch}{1.08}
    \begin{tabular}{@{}p{0.46\columnwidth}p{0.46\columnwidth}@{}}
        \toprule
        \textbf{Component} & \textbf{Setting} \\
        \midrule
        Visual observation & $224\times224$ wrist RGB \\
        Visual encoder & ViT-S/8, trained from scratch \\
        Visual / tactile token & 384-D / 20-D \\
        Fused representation & 404-D \\
        Prediction horizon & 16 steps \\
        Action per step & 9-D wrist + 22-D hand \\
        Diffusion U-Net widths & 256, 512, 1024 \\
        Tactile head & $404\rightarrow512\rightarrow320$, Mish \\
        Training diffusion steps & 50 \\
        DDIM reverse steps & 16 \\
        Executed waypoints & First 8 of each 16-step chunk \\
        Human--robot sampling & Equal probability mass \\
        Tactile-loss weight $\lambda$ & 0.2 \\
        Optimizer & AdamW \\
        Batch size / epochs & 250 / 300 \\
        \bottomrule
    \end{tabular}
\end{table}

The action loss first averages errors within the wrist and hand groups and then weights the two group means equally. This prevents the 22-D finger target from dominating the 9-D wrist target solely because it contains more coordinates. The tactile head regresses the complete $16\times20$ future trajectory from the noise-free interaction representation. Its loss updates both the head and the shared visual encoder, but its prediction is not passed into the diffusion trajectory or the controller.

\section{Rollout and Evaluation Details}
\label{app:evaluation}

\begin{figure*}[!t]
    \centering
    \includegraphics[width=0.88\textwidth]{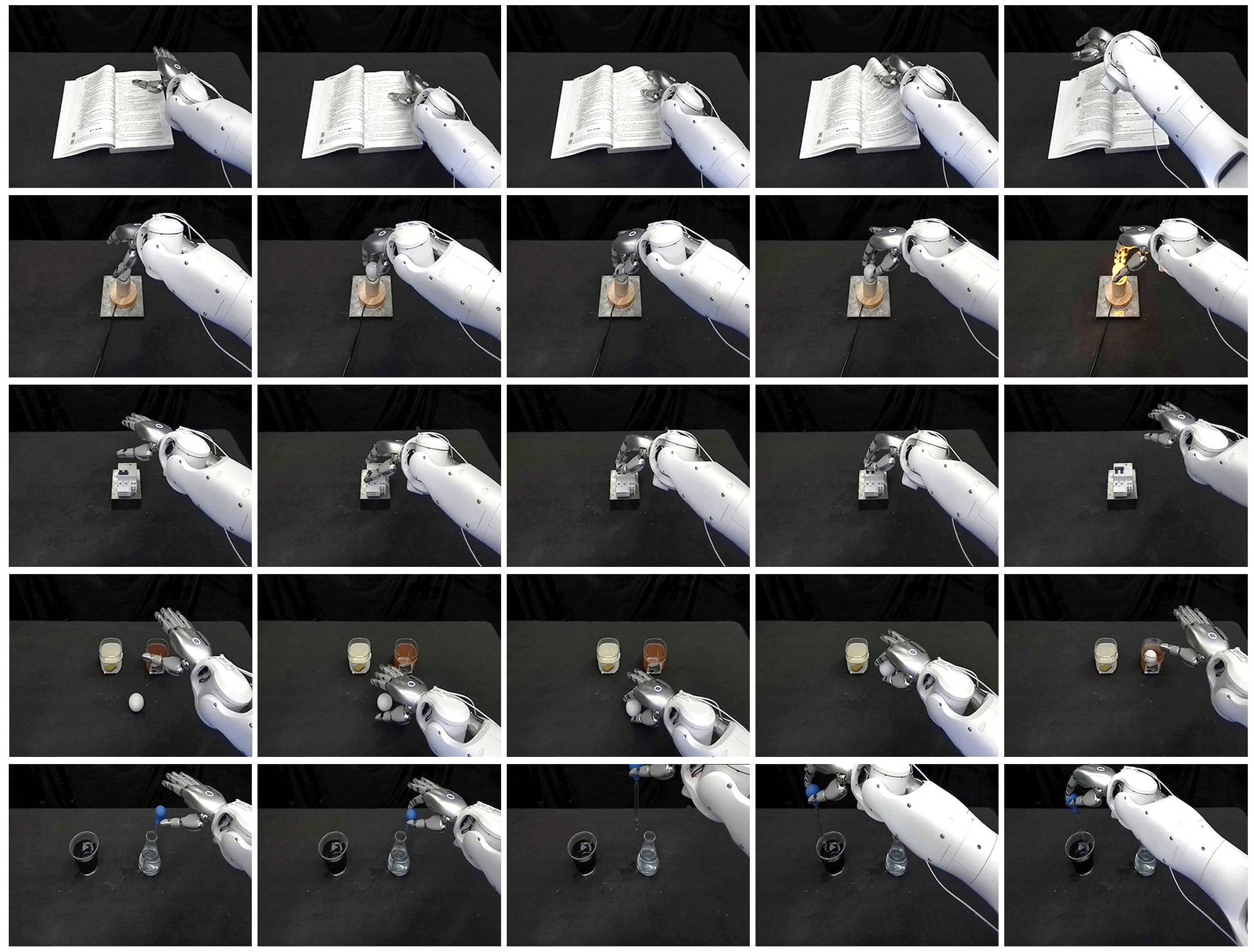}
    \caption{\textbf{Real-Robot Rollout Trajectories.} Representative successful executions of UVTA on five contact-rich tasks. From top to bottom: Flip Page, Screw Light Bulb, Toggle Switch, Ball Classification, and Liquid Transfer with a Dropper. Each row shows an execution sequence, with time progressing from left to right.}
    \label{fig:results}
\end{figure*}

\begin{table*}[!t]
    \centering
    \caption{\textbf{Task-Specific Evaluation Criteria.} A rollout receives the highest score associated with a milestone it achieves; scores are not summed across milestones. A rollout that achieves none of the listed milestones receives 0. A score of 1.0 denotes full task success.}
    \label{tab:success_criteria}
    \small
    \setlength{\tabcolsep}{5pt}
    \renewcommand{\arraystretch}{1.14}
    \begin{tabular}{@{}p{0.21\textwidth}p{\dimexpr0.79\textwidth-10pt\relax}@{}}
        \toprule
        \textbf{Task} & \textbf{Milestones and scores} \\
        \midrule
        Flip Page & \textbf{0.3}: separate and lift the page edge; \textbf{0.7}: turn multiple pages together; \textbf{1.0}: turn exactly one page. \\
        \addlinespace[3pt]
        Screw Light Bulb & \textbf{0.2}: establish stable contact with the bulb; \textbf{0.5}: rotate the bulb through one full turn; \textbf{1.0}: complete multiple turns until the bulb remains lit. \\
        \addlinespace[3pt]
        Toggle Switch & \textbf{0.2}: establish stable contact; \textbf{0.6}: partially actuate the switch; \textbf{1.0}: push the switch fully to its on position. \\
        \addlinespace[3pt]
        Ball Classification & \textbf{0.2}: securely grasp the ball; \textbf{0.6}: place it in a container; \textbf{1.0}: place it in the correct container for its material. \\
        \addlinespace[3pt]
        Liquid Transfer & \textbf{0.2}: establish stable contact with the dropper; \textbf{0.5}: draw liquid into the dropper; \textbf{0.7}: lift and transport the filled dropper; \textbf{1.0}: dispense the liquid into the target container. \\
        \bottomrule
    \end{tabular}
\end{table*}

\subsection{Closed-Loop Execution}
At every policy update, UVTA consumes the current wrist image and tactile token, samples a 16-step robot action chunk with 16 DDIM reverse steps, and denormalizes the chunk with the robot statistics. Relative wrist predictions are anchored to the live wrist pose, while finger predictions are interpreted as absolute joint targets. The controller executes the first eight wrist and hand waypoints, then acquires a new visual-tactile observation and replans. This receding-horizon procedure preserves feedback during contact while avoiding a full diffusion sample at every low-level control step.

\subsection{Stage-Wise Success Rate}
Each method and ablation is evaluated in $N=10$ real-robot rollouts per task, with randomized object positions and heights. Table~\ref{tab:success_criteria} lists the complete scoring criteria, and Fig.~\ref{fig:results} illustrates representative successful rollouts. The reported success rate is a stage-wise score that gives partial credit for task progress, rather than the fraction of fully successful trials.

For task $q$ and rollout $j$, let $w_{q,m}$ be the score assigned to milestone $m$, and let $C_{q,j,m}$ indicate whether that milestone is achieved. We assign a single rollout score
\begin{equation}
    s_{q,j}=\max\!\left(\{0\}\cup
        \{w_{q,m}\mid C_{q,j,m}=1\}\right).
    \label{eq:rollout_score}
\end{equation}
Thus, a rollout receives 0 if it achieves no listed milestone and 1 only if it fully completes the task. Milestone scores are absolute values, not additive rewards. For example, turning multiple pages receives 0.7, while turning exactly one page receives 1.0.

The task-level percentage and the five-task average are
\begin{equation}
    \mathrm{SR}_{q}=\frac{100}{N}\sum_{j=1}^{N}s_{q,j},
    \qquad
    \overline{\mathrm{SR}}=\frac{1}{5}\sum_{q=1}^{5}\mathrm{SR}_{q}.
    \label{eq:success_rate}
\end{equation}
The final column of Table~\ref{tab:main_results} reports this unweighted five-task mean, rounded to the nearest integer percentage. These scores should not be interpreted as binary completion rates.

\subsection{Human-Data Scaling Protocol}
For the human-data scaling study, the robot dataset is fixed at 150 demonstrations per task while the human subset is increased from 0 to 250, 500, 750, and 1,000 demonstrations. The study covers Flip Page, Screw Light Bulb, and Toggle Switch, and uses the same policy architecture, training objective, and evaluation procedure at every data scale.
Its average is computed over these three tasks using the same stage-wise scoring rule.

\newpage
\section{Limitations and Future Work}
\label{app:limitations}

\textbf{Fingertip-only sensing.}
The current tactile motion-capture system measures contact only at the fingertips. This design is motivated by an unresolved sensing limitation: whole-hand tactile gloves available for our data-collection setting cannot yet reliably suppress motion-induced noise during large-range hand movements. Such artifacts make it difficult to distinguish object-contact signals from signals caused by the motion of the glove itself. We therefore restrict sensing to the fingertips to obtain more reliable contact supervision. This choice leaves palm and dorsal-hand contacts unobserved and limits the interaction patterns represented in the dataset. Extending coverage will require improved sensor integration and compensation for motion-induced artifacts.

\textbf{Data and model scale.}
The experiments use five laboratory tasks, a single robot platform, task-specific policies, at most 1,000 human demonstrations per task, and one model scale. Consequently, the present results do not establish how UVTA behaves with substantially larger and more diverse human datasets, larger models, or new robot embodiments. Future work will scale both data and model capacity and investigate multi-task and cross-platform policies alongside more extensive tactile coverage.

\end{document}